\documentclass[runningheads]{llncs}
\usepackage[T1]{fontenc}
\usepackage{graphicx}
\usepackage{multirow} 
\usepackage{makecell} 
\usepackage[table]{xcolor}  
\usepackage{colortbl}   
\usepackage{amsmath} 
\usepackage{marvosym} 
\usepackage[colorlinks=true, linkcolor=black, citecolor=black, urlcolor=magenta, filecolor=black]{hyperref}  

\begin{document}

\title{TractCloud: Registration-free Tractography Parcellation with a Novel Local-global Streamline Point Cloud Representation\thanks{This work was supported by the following NIH grants: R01MH125860, R01MH119222, R01MH132610, R01MH074794, and R01NS125781.}}
\titlerunning{TractCloud: Registration-free Tractography Parcellation}

\author{Tengfei Xue\inst{1,2} \and
Yuqian Chen\inst{1,2} \and
Chaoyi Zhang\inst{2} \and
Alexandra J. Golby\inst{1} \and
Nikos Makris\inst{1} \and
Yogesh Rathi\inst{1} \and
Weidong Cai\inst{2} \and
Fan Zhang\inst{1}$^($\textsuperscript{\Letter}$^)$ \and
Lauren J. O’Donnell\inst{1}$^($\textsuperscript{\Letter}$^)$
}
\authorrunning{T. Xue et al.}
\institute{Harvard Medical School, MA, USA \\ \email{zhangfanmark@gmail.com; odonnell@bwh.harvard.edu} \and The University of Sydney, NSW, Australia}

%
%

%
%
\maketitle              
\begin{abstract}
Diffusion MRI tractography parcellation classifies streamlines into anatomical fiber tracts to enable quantification and visualization for clinical and scientific applications. Current tractography parcellation methods rely heavily  on registration, but registration inaccuracies can affect parcellation and the computational cost of registration is high for large-scale datasets. Recently, deep-learning-based methods have been proposed for tractography parcellation using various types of representations for streamlines. However, these methods only focus on the information from a single streamline, ignoring geometric relationships between the streamlines in the brain. We propose \textit{TractCloud}, a registration-free framework that performs whole-brain tractography parcellation directly in individual subject space. We propose a novel, learnable, local-global streamline representation that leverages information from neighboring and whole-brain streamlines to describe the local anatomy and global pose of the brain. We train our framework on a large-scale labeled tractography dataset, which we augment by applying synthetic transforms including rotation, scaling, and translations. We test our framework on five independently acquired datasets across populations and health conditions. TractCloud significantly outperforms several state-of-the-art methods on all testing datasets. TractCloud achieves efficient and consistent whole-brain white matter parcellation across the lifespan (from neonates to elderly subjects, including brain tumor patients) without the need for registration. The robustness and high inference speed of TractCloud make it suitable for large-scale tractography data analysis. Our project page is available at \href{https://tractcloud.github.io/}{https://tractcloud.github.io/}. 

\end{abstract}
\section{Introduction}
\begin{figure}[t] 
\includegraphics[width=\textwidth]{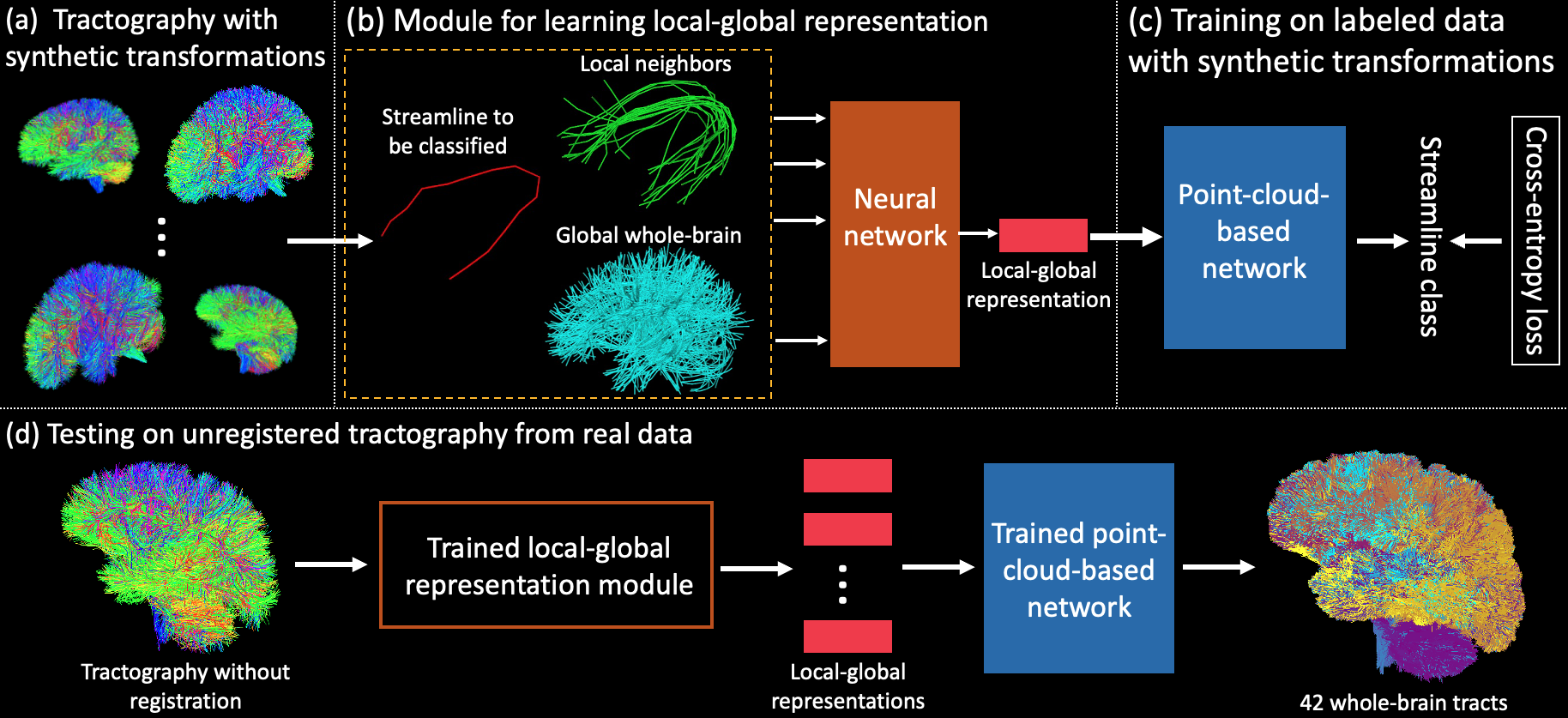}
\caption{TractCloud framework overview: (a) tractography with synthetic transformations, (b) module for learning local-global representation, (c) training on labeled data with synthetic transformations, (d) testing on unregistered tractography from real data.} 
\label{fig_tractcloud}
\end{figure}

Diffusion MRI (dMRI) tractography is the only non-invasive method capable of mapping the complex white matter (WM) connections within the brain \cite{Basser2000-tu}. Tractography parcellation \cite{Zhang2018-jx,Garyfallidis2018-hn,Roman2022-pr} classifies the vast numbers of streamlines resulting from whole-brain tractography to enable visualization and quantification of the brain’s WM connections. (Here a streamline is defined as a set of ordered points in 3D space resulting from tractography \cite{Zhang2022-sj}). In recent years, deep-learning-based methods have been proposed for tractography parcellation \cite{Gupta2017-co,Xu2019-aa,Zhang2020-vm,Chen2021-dm,Xue2023-qc,Kumaralingam2022-ze,Wasserthal2018-if,Liu2022-vk,Wang2022-li,Legarreta2022-hf,Chen2023-fv,Xu2023-zi}, of which many methods are designed to classify streamlines  \cite{Xu2019-aa,Zhang2020-vm,Legarreta2021-ah,Chen2021-dm,Kumaralingam2022-ze,Legarreta2022-hf,Xue2023-qc}. However, multiple challenges exist when using streamline data as deep network input. One well-known challenge is that streamlines can be equivalently represented in forward or reverse order \cite{Garyfallidis2012-rs,Xue2023-qc}, complicating their direct representation as vectors \cite{Chen2021-dm} or images \cite{Zhang2020-vm}. Another challenge is that the geometric relationships between the streamlines in the brain have previously been ignored: existing parcellation methods \cite{Xu2019-aa,Zhang2020-vm,Legarreta2021-ah,Chen2021-dm,Kumaralingam2022-ze,Xue2023-qc,Legarreta2022-hf} train and classify each streamline independently. Finally, computational cost can pose a challenge for the parcellation of large tractography datasets that can include thousands of subjects with millions of streamlines per subject. \newline
\indent In this work, we propose a novel point-cloud-based strategy that leverages neighboring and whole-brain streamline information to learn local-global streamline representations. Point clouds have been shown to be efficient and effective representations for streamlines \cite{Astolfi2020-jf,Chen2022-uz,Kumaralingam2022-ze,Xue2023-qc,Chen2023-bn,Li2022-bx} in applications such as tractography filtering \cite{Astolfi2020-jf}, clustering \cite{Chen2021-dm}, and parcellation \cite{Xue2022-yz,Kumaralingam2022-ze,Xue2023-qc,Li2022-bx}. One benefit of using point clouds is that streamlines with equivalent forward and reverse point orders (e.g., from cortex to brainstem or vice versa) can be represented equally. However, these existing methods focus on a single streamline (one point cloud) and ignore other streamlines (other point clouds) in the same brain that may provide important complementary information useful for tractography parcellation. In computer vision, point clouds are commonly used to describe scenes and objects (e.g., cars, tables, airplanes, etc.). However, point cloud segmentation methods from computer vision, which assign labels to points, cannot translate directly to the tractography field, where the task of interest is to label entire streamlines. Computer vision studies \cite{Qi2017-ju,Wang2019-az,Yan2020-rj,Xiang2021-cg,Zhao2021-bv,Yu2021-ti,Ma2022-gg} have shown that point interactions \textit{within one point cloud} can yield more effective features for downstream tasks. However, in tractography parcellation we are interested in the relationship \textit{between multiple point clouds (streamlines)} in the brain. These other streamlines can provide detailed information about the local WM geometry surrounding the streamline to be classified, as well as global information about the location and pose of the brain that can reduce the need for image registration. \newline
\indent Affine or even nonrigid registration is needed for current tractography parcellation methods~\cite{Garyfallidis2015-ox,Zhang2018-jx,Roman2022-pr}. Recently, registration-free techniques have been proposed for tractography parcellation to handle computational challenges resulting from large inter-subject variability and to increase robustness to image registration inaccuracies \cite{Liu2019-jj,Siless2020-kx}. Avoiding image registration can also reduce computational time and cost when processing very large tractography datasets with thousands of subjects. While other registration-free tractography parcellation techniques require Freesurfer input \cite{Siless2020-kx} or work with rigidly MNI-aligned Human Connectome Project data \cite{Liu2019-jj}, our method can directly parcellate tractography in individual subject space. \newline
\indent In this study, we propose \textit{TractCloud}, a registration-free tractography parcellation framework, as illustrated in Fig.~\ref{fig_tractcloud}. This paper has three main contributions. First, we propose a novel, learnable, local-global streamline representation that leverages information from neighboring and whole-brain streamlines to describe the local anatomy and global pose of the brain. Second, we leverage a training strategy using synthetic transformations of labeled tractography data to enable registration-free parcellation at the inference stage. Third, we implement our framework using two compared point cloud networks and demonstrate fast, registration-free, whole-brain tractography parcellation across the lifespan.

\section{Methods}
\subsection{Training and Testing Datasets}
We utilized a high-quality and large-scale dataset of 1 million labeled streamlines for model training and validation. The dataset was obtained from a WM tractography atlas \cite{Zhang2018-jx} that was curated and annotated by a neuroanatomist. The atlas was derived from 100 registered tractography of young healthy adults in the Human Connectome Project (HCP) \cite{Van_Essen2013-ne}. The training data includes 43 tract classes: 42 anatomically meaningful tracts from the whole brain and one tract category of “other streamlines,” including, most importantly, anatomically implausible outlier streamlines. On average, the 42 anatomical tracts have 2539 streamlines with a standard deviation of 2693 streamlines.

For evaluation, we used a total of 120 subjects from four public datasets and one private dataset. These five datasets were independently acquired with different imaging protocols across ages and health conditions. (1) developing HCP (dHCP) \cite{Edwards2022-vr}: 20 neonates (1 to 27 days); (2) Adolescent Brain Cognitive Development (ABCD) dataset \cite{Volkow2018-og}: 25 adolescents (9 to 11 years); (3) HCP dataset \cite{Van_Essen2013-ne}: 25 young healthy adults (22 to 35 years, subjects not part of the training atlas); (4) Parkinson’s Progression Markers Initiative (PPMI) dataset \cite{Parkinson_Progression_Marker_Initiative2011-oz}: 25 older adults (51 to 75 years), including Parkinson's disease (PD) patients and healthy individuals; (5) Brain Tumor Patient (BTP) dataset: dMRI data from 25 brain tumor patients (28 to 70 years) were acquired at Brigham and Women’s Hospital. dMRI acquisition parameters of datasets are shown in Supplementary Table S1. The two-tensor Unscented Kalman Filter (UKF) \cite{Malcolm2010-mk,Reddy2016-ko,Norton2017-dx} method, which is consistent across ages, health conditions, and image acquisitions \cite{Zhang2018-jx}, was utilized to create whole-brain tractography for all subjects across the datasets mentioned above.

\subsection{TractCloud Framework}

\subsubsection{Synthetic Transform Data Augmentation.}
To enable tractography parcellation without registration, we augmented the training data by applying synthetic transform-based augmentation (STA) including rotation, scaling, and translations. These transformations have been used in voxel-based WM segmentation \cite{Wasserthal2018-if}, but no study has applied these transformations to study tractography, to our knowledge. In detail, we applied 30 random transformations to each subject tractography in the  training dataset to obtain 3000 transformed subjects and 30 million streamlines. Transformations included: rotation from -45 to 45 degrees along the left-right axis, from -10 to 10 degrees along the anterior-posterior axis, and from -10 to 10 degrees along the superior-inferior axis; translation from -50 to 50 mm along all three axes; scaling from -45\% to 5\% along all three axes. These transformations were selected based on typical differences between subjects due to variability in brain anatomy and volume, head position, and image acquisition protocol. Many methods are capable of tractography parcellation after affine registration \cite{Garyfallidis2018-hn,Zhang2018-jx}; therefore, with STA applied to the training dataset, our framework has the potential for registration-free parcellation. 

\begin{figure}[t] 
\includegraphics[width=\textwidth]{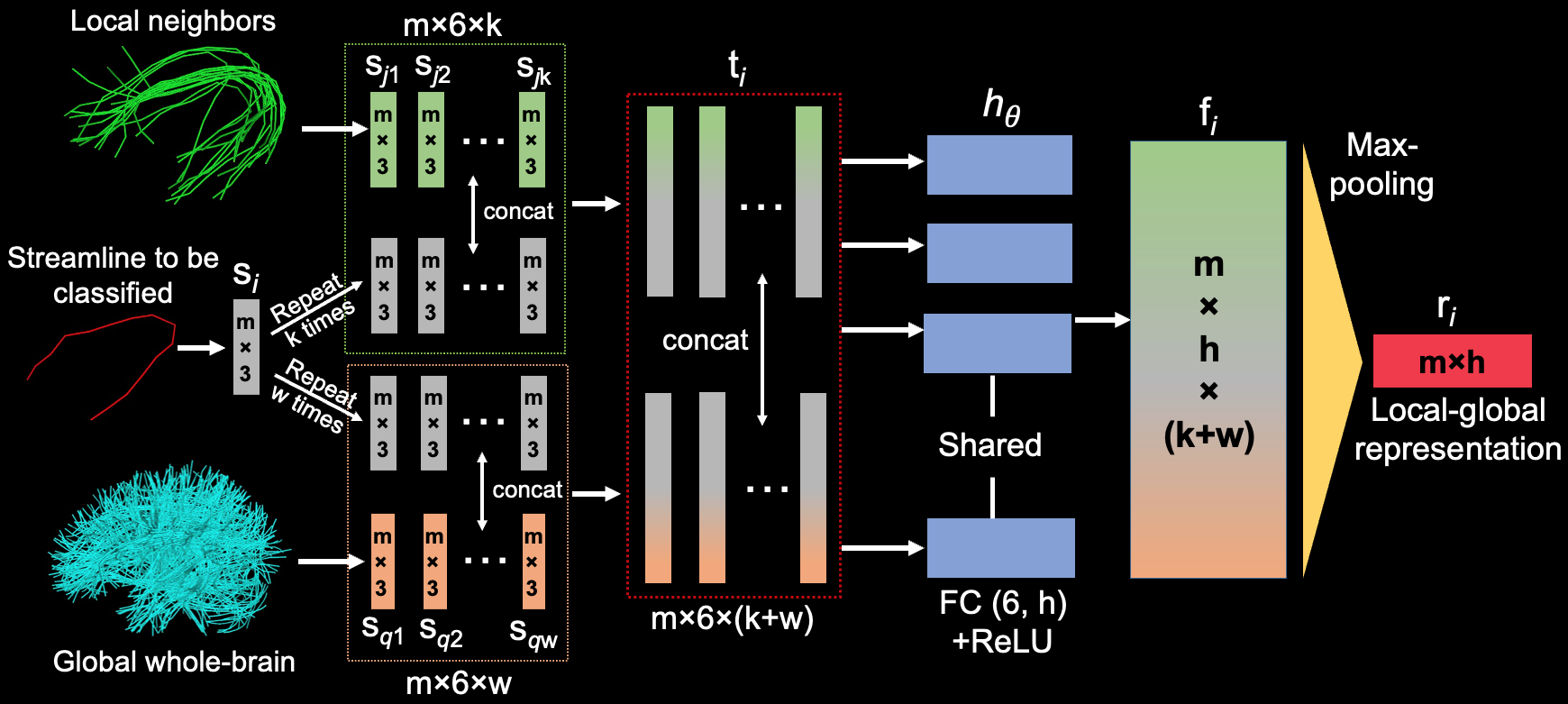}
\caption{The proposed module for learning local-global representation.} 
\label{fig_localglobal}
\end{figure}

\subsubsection{Module for Local-global Streamline Representation Learning.}
We propose a module (Fig.~\ref{fig_localglobal}) to learn the proposed local-global representation, which benefits from information about the anatomy of the neighboring WM and the overall pose of the brain. We construct the input for the learning module by concatenating the coordinates of the original streamline (the one to be classified), its local neighbor streamlines, and global whole-brain streamlines. In detail, assume a brain has $\emph{n}$ streamlines, denoted by $ \emph{S} = \{s_{1}, s_{2}, \ldots, s_{n}\}$, $ s_{i} \in \mathcal{R}^{m\times 3}$, where 3 is the dimensionality of the point coordinates and $\emph{m}$ is the number of points for each streamline ($\emph{m}$=15 as in \cite{Zhang2018-jx,Zhang2020-vm}). For streamline $s_{i}$, we obtain a set of $\emph{k}$ nearest streamlines, $\emph{local}(s_{i}) = \{s_{j1}, s_{j2},\ldots, s_{jk}\}$, using a pairwise streamline distance \cite{Garyfallidis2012-rs}. From the whole brain, we also randomly select a set of $\emph{w}$ streamlines, $\emph{global}(s_{i}) = \{s_{q1}, s_{q2},\ldots, s_{qw}\}$. Then $s_{i}$, $\emph{local}(s_{i})$, and $\emph{global}(s_{i})$ are concatenated as shown in Fig.~\ref{fig_localglobal} to obtain the input of the module, $ t_{i} \in \mathcal{R}^{m \times 6 \times (k+w)}$. The proposed module begins with a shared fully connected (FC) layer with ReLU activation function ($h_{\Theta}$): $f_{i} = h_{\Theta} (t_{i})$, $f_{i} \in \mathcal{R}^{m \times h \times (k+w)}$, where $\emph{h}$ is the output dimension of $h_{\Theta}$ ($\emph{h}$=64 \cite{Charles2017-nv,Qi2017-ju,Wang2019-az}). Finally, the local-global representation $r_{i}$ is obtained through max-pooling $r_{i}$ = pool($f_{i}$) $\in \mathcal{R}^{m \times h}$. The network is trained in an end-to-end fashion where the local-global representation $r_{i}$ is learned during training of the overall point-cloud-based classification network.

\subsubsection{Network Structure for Streamline Classification.}
The local-global representation  learning module can replace the first layer or module of typical point-cloud-based networks \cite{Charles2017-nv,Qi2017-ju,Wang2019-az,Zhao2021-bv}. Here, we explore two widely used networks: PointNet \cite{Charles2017-nv} and Dynamic Graph Convolutional Neural Network (DGCNN) \cite{Wang2019-az}. PointNet (see Fig. S1 for network details) encodes point-wise features individually, but DGCNN (see Fig. S2 for network details) encodes point-wise features by interacting with other points on a streamline. Both PointNet and DGCNN then aggregate features of all points through pooling to get a single streamline descriptor, which is input into fully connected layers for classification.

\subsection{Implementation Details}
To learn $r_{i}$, we used 20 local streamlines (selected from 10, 20, 50, 100) and 500 global streamlines (selected from 100, 300, 500, 1000). Our framework was trained with the Adam optimizer with a learning rate of 0.001 using cross-entropy loss. The epoch was 20, and the batch size was 1024. Training of our registration-free framework (TractCloud\textsubscript{\textit{reg-free}}) with the large STA dataset took about 22 hours and 10.9 GB GPU memory with Pytorch (v1.13) on an NVIDIA RTX A5000 GPU machine. For training and inference using TractCloud\textsubscript{\textit{reg-free}}, tractography was centered at the mass center of the training atlas. TractCloud\textsubscript{\textit{reg-free}} directly annotates tract labels at the inference stage without requiring the registration of an atlas.

\section{Experiments and Results}
\subsection{Performance on the Labeled Atlas Dataset}
We evaluated our method on the original labeled training dataset (registered and aligned) and its synthetic transform augmented (STA) data (unregistered and unaligned). We divided both the original and STA data into train/validation/test sets with the distribution of 70\%/10\%/20\% by subjects (such that all streamlines from an individual subject were placed into only one set, either train or validation or test). For experimental comparison, we included two deep-learning-based state-of-the-art (SOTA) tractography parcellation methods:  DCNN++ \cite{Xu2019-aa} and DeepWMA \cite{Zhang2020-vm}. They were both designed to perform deep WM parcellation using CNNs, with streamline spatial coordinate features as input. We trained the networks based on the recommended settings in their papers and code. Two widely used point-cloud-based networks (PointNet \cite{Charles2017-nv} and DGCNN \cite{Wang2019-az}), with a single streamline as input, were included as baseline methods. To evaluate the effectiveness of the local-global representation in TractCloud, we performed experiments using only local neighbor features (PointNet\textsubscript{\textit{+loc}} and DGCNN\textsubscript{\textit{+loc}}) and both local neighbor and whole-brain global features (PointNet\textsubscript{\textit{+loc+glo}} and DGCNN\textsubscript{\textit{+loc+glo}}). For all methods, we report two metrics (accuracy and macro F1) that are widely used for tractography parcellation \cite{Ngattai_Lam2018-ne,Liu2019-jj,Xu2019-aa,Zhang2020-vm,Xue2023-qc}. The accuracy is reported as the overall accuracy of streamline classification, and the macro F1 score is reported as the mean across 43 tract classes (Table~\ref{tab_res_labeled}). \newline
\begin{table}[t]
\caption{Results on the labeled training dataset with and without synthetic transformations. Bold and italic text indicates the best-performing method and the second-best-performing method, respectively. Abbreviations: Orig - Original, Acc - Accuracy.}
\centering
\begin{tabular}{|c|c|c|c|c|c|c|c|c|c|}
\hline
\multicolumn{2}{|c|}{Feature} &
  \multicolumn{4}{c|}{\begin{tabular}[c]{@{}c@{}}Single Streamline \end{tabular}} &
  \multicolumn{2}{c|}{\begin{tabular}[c]{@{}c@{}}Local \end{tabular}} &
  \multicolumn{2}{c|}{\begin{tabular}[c]{@{}c@{}}Local + Global \end{tabular}} \\
\hline
\multicolumn{2}{|c|}{} &
  \multicolumn{2}{c|}{\begin{tabular}[c]{@{}c@{}}SOTA \\ Methods\end{tabular}} &
  \multicolumn{2}{c|}{\begin{tabular}[c]{@{}c@{}}Point Cloud \\ Networks\end{tabular}} &
  \multicolumn{4}{c|}{\begin{tabular}[c]{@{}c@{}}TractCloud \\ Effectiveness Study\end{tabular}} \\
  \cline{3-10}
  \multicolumn{2}{|c|}{\multirow{-3}{*}{\begin{tabular}[c]{@{}c@{}}Data \\ \& \\ Metric\end{tabular}}} &
  \begin{tabular}[c]{@{}l@{}}Deep\\ WMA\end{tabular} &
  \begin{tabular}[c]{@{}l@{}}DCNN++\end{tabular} &
  \begin{tabular}[c]{@{}l@{}}PointNet\end{tabular} &
  \begin{tabular}[c]{@{}l@{}}DGCNN\end{tabular} &
    \begin{tabular}[c]{@{}l@{}}PointNet \\~~~\textsubscript{\textit{+loc}}\end{tabular} &
  \begin{tabular}[c]{@{}l@{}}DGCNN \\~~~\textsubscript{\textit{+loc}} \end{tabular} &
  \begin{tabular}[c]{@{}l@{}}PointNet \\ \textsubscript{\textit{+loc+glo}} \end{tabular} &
  \begin{tabular}[c]{@{}l@{}}DGCNN \\ \textsubscript{\textit{+loc+glo}}\end{tabular} \\
\hline
 &
  Acc &
  90.29 &
  91.26 &
  91.36 &
  91.85 &
  91.51 &
  91.91 &
  \textbf{92.28} &
  \textit{91.99} \\
\cline{2-10}  
\multirow{-2}{*}{\begin{tabular}[c]{@{}l@{}}Orig\\ data\end{tabular}} &
  F1 &
  88.12 &
  89.14 &
  89.12 &
  89.78 &
  89.25 &
  90.03 &
  \textbf{90.36} &
  \textit{90.10} \\
\hline
 &
  Acc &  
  82.35 &
  84.14 &
  81.83 &
  83.70 &
  86.56 &
  87.14 &
  \textit{91.57} &
  \textbf{91.69} 
\\
\cline{2-10}  
\multirow{-2}{*}{\begin{tabular}[c]{@{}l@{}}STA\\ data\end{tabular}} &
  F1 &
  76.55 &
  79.16 &
  75.89 &
  78.55 &
  82.08 &
  82.95 &
  \textit{89.40} &
  \textbf{89.65}  \\
\hline
\end{tabular}
\label{tab_res_labeled}
\end{table}
\indent Table~\ref{tab_res_labeled} shows that the TractCloud framework achieves the best performance on data with and without synthetic transformations (STA). Especially on STA data, TractCloud yields a large improvement in accuracy (up to 9.9\%) and F1 (up to 13.8\%), compared to PointNet and DGCNN baselines as well as SOTA methods. In addition, including local (PointNet\textsubscript{\textit{+loc}} and DGCNN\textsubscript{\textit{+loc}}) and global (PointNet\textsubscript{\textit{+loc+glo}} and DGCNN\textsubscript{\textit{+loc+glo}}) features both improve the performance compared to baselines (PointNet and DGCNN) with a single streamline as input. This demonstrates the effectiveness of our local-global representation.

\subsection{Performance on the Independently Acquired Testing Datasets}
We performed experiments on five independently acquired, unlabeled testing datasets (dHCP, ABCD, HCP, PPMI, BTP) to evaluate the robustness and generalization ability of our TractCloud\textsubscript{\textit{reg-free}} framework on unseen and unregistered data. All compared SOTA methods (DeepWMA, DCNN++) and TractCloud\textsubscript{\textit{regist}} were tested on registered tractography,  and only  TractCloud\textsubscript{\textit{reg-free}} was tested on unregistered tractography. Tractography was registered to the space of the training atlas using an affine transform produced by registering the baseline (b=0) image of each subject to the atlas population mean T2 image using 3D Slicer \cite{Fedorov2012-ma}. For each method, we quantified the tract identification rate (TIR) and calculated the tract-to-atlas distance (TAD), and statistical significance tests were performed for results of TIR and TAD (Table~\ref{tab_res_test}).  TIR measures if the tract is identified successfully when labels are not available \cite{Zhang2018-jx,Zhang2020-vm,Chen2021-dm}. Here, we chose 50 as the minimum number of streamlines for a tract to be considered as identified (The threshold of 50 is more strict than 10 or 20 in \cite{Zhang2018-jx,Zhang2020-vm,Chen2021-dm}). As a complementary metric for TIR, TDA measures the geometric similarity between identified tracts and corresponding tracts from the training atlas. For each testing subject's tract, we calculated the streamline-specific minimum average direct-flip distance  \cite{Garyfallidis2012-rs,Zhang2018-jx,Chen2021-dm} to the atlas tract and then computed the average across subjects and tracts to obtain TDA. We also recorded the computation time for tractography parcellation for every method (Table~\ref{tab_res_test}). The computation time was tested on a Linux workstation with an NVIDIA RTX A4000 GPU using tractography (0.28 million streamlines) from a randomly selected subject. To evaluate if differences in result values between our registration-free method (TractCloud\textsubscript{\textit{reg-free}}) and other methods are significant, we implemented a repeated measure ANOVA test for all methods across subjects, and then we performed multiple paired Student’s $t$-tests between TractCloud\textsubscript{\textit{reg-free}} method and each compared method. In addition, in order to evaluate how well our framework can perform without registration, we converted identified tracts into volume space and calculated the spatial overlap (weighted Dice) \cite{Cousineau2017-nf,Zhang2019-bl} between results of TractCloud\textsubscript{\textit{regist}} and TractCloud\textsubscript{\textit{reg-free}} (Table~\ref{tab_res_overlap}). Furthermore, we also provide a visualization of identified tracts in an example individual subject for every dataset across methods (Fig.~\ref{fig_localglobal}). \newline
\begin{table}[t]
\caption{Results of tract identification rate (TIR) and tract distance to atlas (TDA) on five independently acquired testing datasets as well as computation time on a randomly selected subject. TIR results show no significant differences across methods (ANOVA $p > 0.05$), while TDA results do (ANOVA $p < 1\times10^{-10}$). Asterisks show that the difference between TractCloud\textsubscript{\textit{reg-free}} and other methods is significant using a paired Student’s $t$-test. ($*p < 0.05$, $**p < 0.001$). Abbreviations: TC - TractCloud.}
\begin{tabular}{|l|c|c|c|c|c|l|l|l|l|l|c|}
\hline
\multicolumn{1}{|c|}{\multirow{2}{*}{Method}} &
  \multicolumn{5}{c|}{TIR (\%) $\uparrow$} &
  \multicolumn{5}{c|}{TDA (mm) $\downarrow$} &
  \multirow{2}{*}{\begin{tabular}[c|]{@{}c@{}}Run\\ Time \end{tabular}} \\
\cline{2-11}
\multicolumn{1}{|c|}{} &
  dHCP &
  ABCD &
  HCP &
  PPMI &
  BTP &
  \multicolumn{1}{c|}{dHCP} &
  \multicolumn{1}{c|}{ABCD} &
  \multicolumn{1}{c|}{HCP} &
  \multicolumn{1}{c|}{PPMI} &
  \multicolumn{1}{c|}{BTP} &
   \\
\hline
DeepWMA       & \makecell[c]{98.8 \\ $\pm$1.4} & 100 & 100 & 100 & \makecell[c]{99.9 \\ $\pm$0.5} & \makecell[c]{$6.81^{*}$ \\ $\pm$1.1} & \makecell[c]{$5.66^{**}$ \\ $\pm$0.9} 
               & \makecell[c]{$5.09^{**}$ \\ $\pm$0.7} & \makecell[c]{$5.94^{**}$ \\ $\pm$1.0} & \makecell[c]{$6.24^{*}$ \\ $\pm$1.2} & 113s \\
\hline
DCNN++         & \makecell[c]{98.7 \\ $\pm$1.9} &100 &100 & 100 & \makecell[c]{99.8 \\ $\pm$0.9} & \makecell[c]{$6.90^{*}$ \\ $\pm$1.4} & \makecell[c]{$5.69^{**}$ \\ $\pm$0.9} 
               & \makecell[c]{$5.08^{**}$ \\ $\pm$0.7} & \makecell[c]{$5.95^{**}$ \\ $\pm$0.9} & \makecell[c]{$6.43^{**}$ \\ $\pm$1.8} & 102s \\
\hline
\rowcolor{gray!8}
TC\textsubscript{\textit{regist}}   & \makecell[c]{99.2 \\ $\pm$1.6}  & 100 & 100 & 100 & \makecell[c]{99.9 \\ $\pm$0.5} & \makecell[c]{$6.53^{*}$ \\ $\pm$1.1} & \makecell[c]{$5.60^{**}$ \\ $\pm$0.9} 
                                    & \makecell[c]{$5.06^{**}$ \\ $\pm$0.7} & \makecell[c]{$5.87^{*}$ \\ $\pm$1.0} & \makecell[c]{6.09 \\ $\pm1.1$} & 97s  \\
\hline
\rowcolor{gray!8}
TC\textsubscript{\textit{reg-free}} & \makecell[c]{97.7 \\ $\pm$3.1} & 100 & 100 &100 & \makecell[c]{99.9 \\ $\pm$0.5}   & \makecell[c]{6.71 \\ $\pm$1.3} & \makecell[c]{5.52\\ $\pm$0.8} 
                                    & \makecell[c]{5.02 \\ $\pm$0.6} & \makecell[c]{5.85 \\ $\pm$0.9} & \makecell[c]{6.11 \\ $\pm$1.0} & 58s \\
\hline
\end{tabular}
\label{tab_res_test}
\end{table}
\indent As shown in Table~\ref{tab_res_test}, all methods achieve high TIRs on all datasets, and the TIR metric does not have significant differences across methods. This demonstrates that most tracts can be identified by all methods robustly. However, our registration-free framework (TractCloud\textsubscript{\textit{reg-free}}) obtains significantly lower TDA values (better quality of identified tracts) than all compared methods on ABCD, HCP, and PPMI datasets, where ages of test subjects are from 9 to 75 years old. On the very challenging dHCP (baby brain) dataset, TractCloud\textsubscript{\textit{reg-free}} still significantly outperforms two SOTA methods. Note that TractCloud\textsubscript{\textit{reg-free}} directly works on unregistered tractography from neonate brains (much smaller than adult brains). In the challenging BTP (tumor patients) dataset,  TractCloud\textsubscript{\textit{reg-free}} obtains significantly lower TDA values than SOTA methods and comparable performance to TractCloud\textsubscript{\textit{regist}}. As shown in Table~\ref{tab_res_test}, our registration-free framework is much faster than other compared methods.

\begin{figure}[htbp] 
\includegraphics[width=\textwidth]{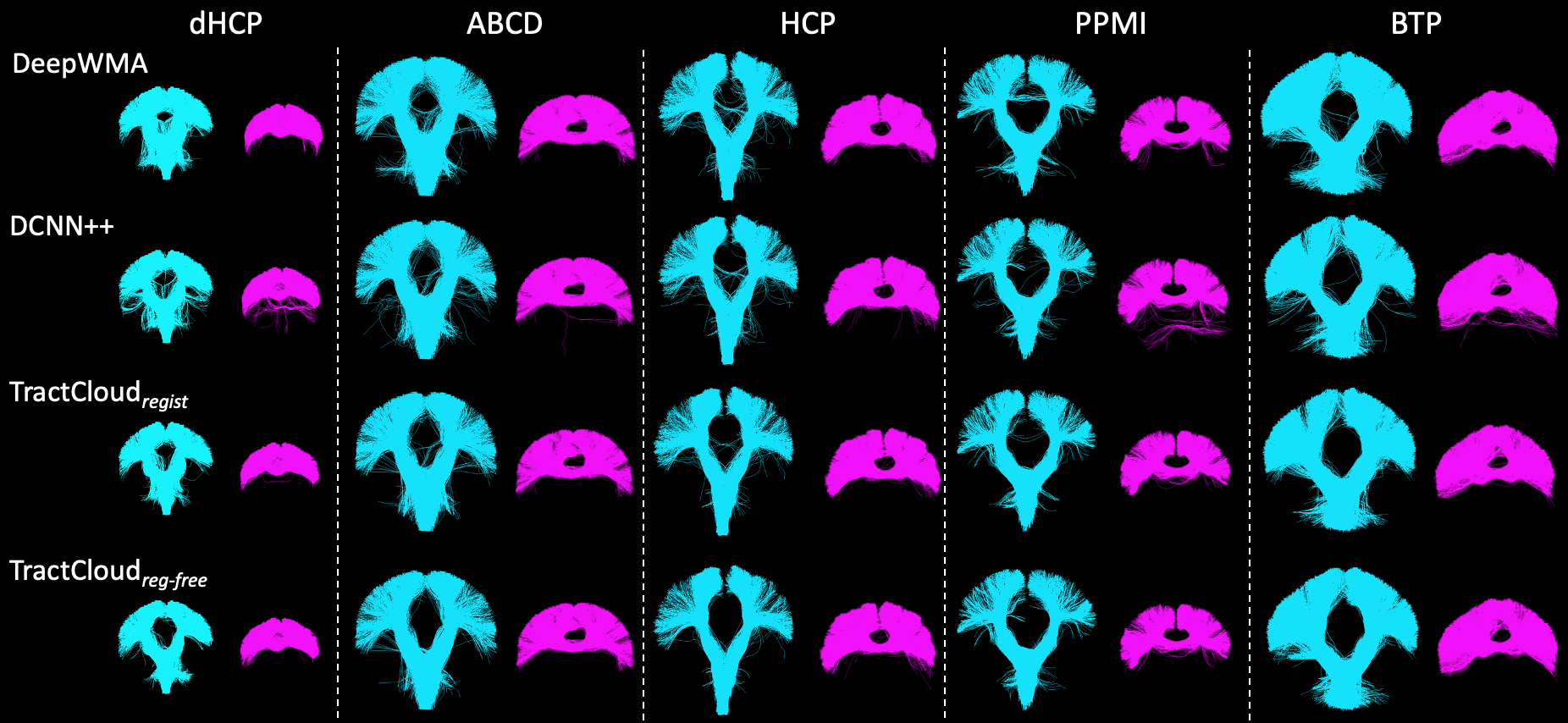}
\caption{Visualization of example tracts (corticospinal tract and corpus callosum IV) from each method, in the subject with the median TDA for each testing dataset. }
\label{fig_visual}
\end{figure}

\begin{table}[t]
\caption{Tract spatial overlap (wDice) between TractCloud\textsubscript{\textit{regist}} and TractCloud\textsubscript{\textit{reg-free}}}
\centering
\begin{tabular}{|c|c|c|c|c|c|}
\hline 
  &  dHCP & ABCD & HCP & PPMI & BTP \\
\hline
 TSO & 0.932$\pm$0.14 & 0.965$\pm$0.04 & 0.980$\pm$0.02 & 0.977$\pm$0.03 & 0.970$\pm$0.05 \\
\hline
\end{tabular}
\label{tab_res_overlap}
\end{table}
The tract spatial overlap (wDice) is over 0.965 on all datasets, except for the challenging dHCP (wDice is 0.932) (Table~\ref{tab_res_overlap}). Overall, our registration-free framework is comparable to (or better than) our framework with registration.

Fig.~\ref{fig_visual} shows visualization results of example tracts. All methods can successfully identify these tracts across datasets. It is visually apparent that the TractCloud\textsubscript{\textit{reg-free}} framework obtains results with fewer outlier streamlines, especially on the challenging dHCP dataset. 

\section{Discussion and Conclusion}
We have demonstrated TractCloud, a registration-free tractography parcellation framework with a novel, learnable, local-global representation of streamlines. Experimental results show that TractCloud can achieve efficient and consistent tractography parcellation results across populations and dMRI acquisitions, with and without registration. The fast inference speed and robust ability to parcellate data in original subject space will allow TractCloud to be useful for analysis of large-scale tractography datasets. Future work can investigate additional data augmentation using local deformations to potentially increase robustness to pathology. Overall, TractCloud demonstrates the feasibility of registration-free tractography parcellation across the lifespan.

\bibliographystyle{splncs04}
\bibliography{refs}
%





\newpage
\section*{Supplementary Material}
\setcounter{table}{0} 
\setcounter{figure}{0} 

\renewcommand{\thetable}{S\arabic{table}}
\begin{table}[htbp]
\caption{dMRI acquisition parameters for five independently acquired testing datasets.}
\centering
\begin{tabular}{|l|l|}
\hline
Dataset & dMRI acquisition parameters  \\ 
\hline
dHCP & \begin{tabular}[c]{@{}l@{}}b = 400/1000/2600 $s/mm^{2}$; \\ 20 volumes with b = 0~$s/mm^{2}$ , 64 volumes with b = 400 $s/mm^{2}$, \\ 88 volumes with b = 1000 $s/mm^{2}$, 128 volumes with b = 2600 $s/mm^{2}$; \\ TE/TR = 90/3800 $ms$;\\ resolution = 1.5x1.5x1.5 $mm^{3}$ \end{tabular} \\ 
\hline
ABCD    & \begin{tabular}[c]{@{}l@{}}b = 3000 $s/mm^{2}$; \\ 1 volume with  b = 0~$s/mm^{2}$, 60 volumes with b = 3000 $s/mm^{2}$; \\ TE/TR = 88/4100 $ms$; \\ resolution = 1.7x1.7x1.7 $mm^{3}$\end{tabular} \\ 
\hline
HCP  & \begin{tabular}[c]{@{}l@{}}b = 3000 $s/mm^{2}$; \\ 18 volumes with b = 0~$s/mm^{2}$, 90 volumes with b = 3000 $s/mm^{2}$; \\ TE/TR = 89/5520 $ms$; \\ resolution = 1.25x1.25x1.25 $mm^{3}$\end{tabular}  \\ 
\hline
PPMI    & \begin{tabular}[c]{@{}l@{}}b = 1000 $s/mm^{2}$; \\ 1 volume with b = 0~$s/mm^{2}$, 64 volumes with b = 1000 $s/mm^{2}$;  \\ TE/TR = 88/7600 $ms$;\\ resolution = 2x2x2 $mm^{3}$\end{tabular}   \\ 
\hline
BTP     & \begin{tabular}[c]{@{}l@{}}b = 2000 $s/mm^{2}$; \\ 1 volume with b = 0~$s/mm^{2}$, 30 volumes with b = 2000 $s/mm^{2}$; \\ TE/TR = 98/12700 $ms$;\\ resolution = 2.2x2.2x2.3 $mm^{3}$\end{tabular} \\ 
\hline
\end{tabular}
\label{tab_dMRIparams}
\end{table}

\newpage

\renewcommand{\thefigure}{S\arabic{figure}}
\begin{figure}[htbp] 
\centering
\includegraphics[width=0.9 \textwidth]{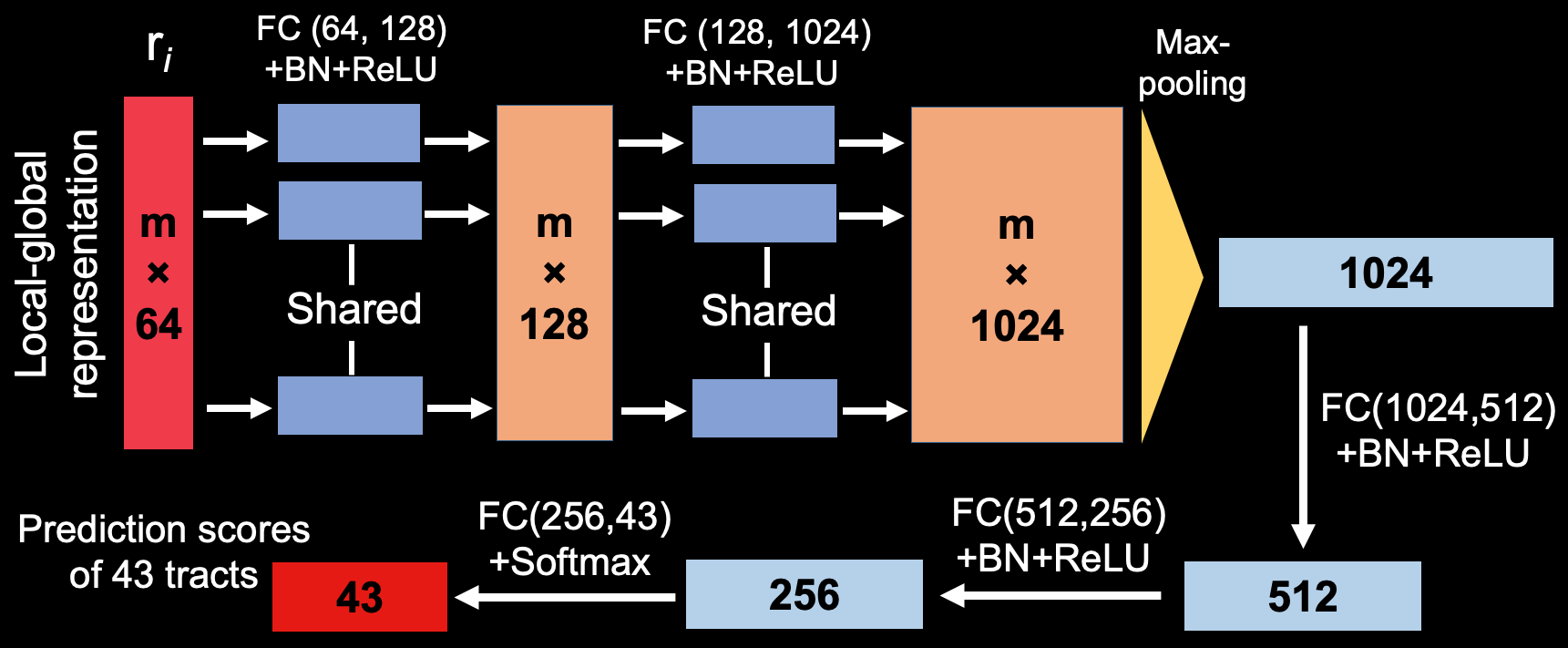}
\caption{The point-cloud-based network architecture of TractCloud using PointNet in our study. $m$ is the number of points on a streamline. Abbreviations: FC, fully connected; BN, batch normalization; ReLU, rectified linear unit. } 
\label{fig_pointnet}
\end{figure}

\renewcommand{\thefigure}{S\arabic{figure}}
\begin{figure}[htbp] 
\centering
\includegraphics[width=\textwidth]{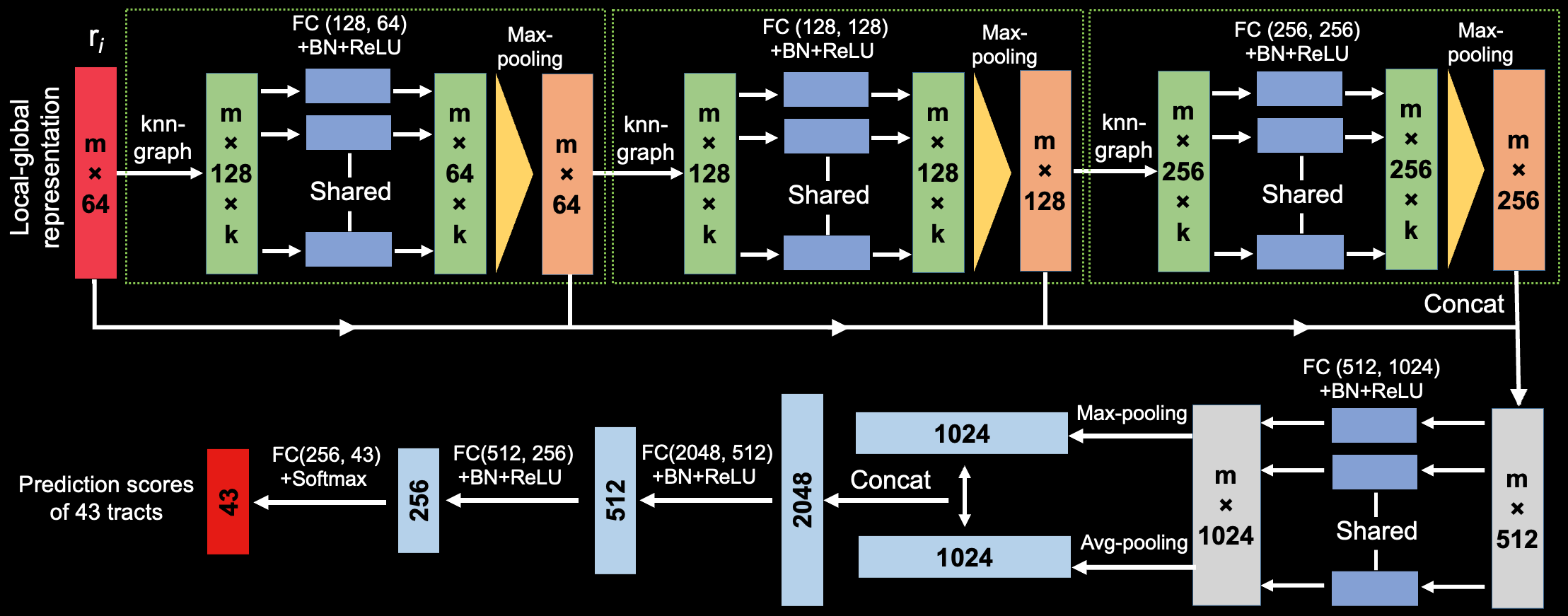}
\caption{The point-cloud-based network architecture of TractCloud using DGCNN in our study. $m$ is the number of points on a streamline. Abbreviations: knn, k-nearest neighbors; FC, fully connected; BN, batch normalization; ReLU, rectified linear unit; Avg, average.} 
\label{fig_dgcnn}
\end{figure}

\end{document}